\documentclass[conference]{IEEEtran}

\usepackage{amsmath,amssymb,amsfonts}
\usepackage{graphicx}
\usepackage{booktabs}
\usepackage{cite}
\usepackage{hyperref}
\usepackage{url}
\usepackage{multirow}
\usepackage{array}
\usepackage{tabularx}

\hypersetup{
  colorlinks=true,
  linkcolor=black,
  citecolor=black,
  urlcolor=blue
}

\begin{document}

\title{Implicit Regularization and Generalization in Overparameterized Neural Networks}

\author{\IEEEauthorblockN{Zeran Johannsen}
\IEEEauthorblockA{ORCID: \href{https://orcid.org/0009-0000-5996-0215}{0009-0000-5996-0215}}}

\maketitle

\begin{abstract}
Modern deep neural networks are frequently trained in regimes where the number of model parameters greatly exceeds the number of training samples. Classical statistical learning theory predicts that such highly overparameterized models should exhibit severe overfitting and poor generalization performance. However, empirical evidence across computer vision, natural language processing, and reinforcement learning consistently demonstrates that large neural networks trained with gradient-based optimization methods often generalize remarkably well. This apparent contradiction has become one of the central theoretical questions in contemporary machine learning research.

Prior theoretical frameworks based on classical capacity measures, such as VC dimension and uniform convergence bounds, struggle to adequately explain why heavily overparameterized neural networks avoid overfitting in practice. While these frameworks predict that increasing model capacity should degrade generalization performance, modern neural networks often show the opposite behavior. This discrepancy has motivated a growing body of research aimed at identifying mechanisms that implicitly constrain the effective capacity of neural networks during training.

This study investigates the role of optimization dynamics and implicit regularization in enabling generalization in overparameterized neural networks through controlled computational experiments. The analysis examines the behavior of stochastic gradient descent (SGD) across batch sizes, the geometry of flat versus sharp minima in the loss landscape via Hessian eigenvalue estimation and weight perturbation analysis, theoretical perspectives derived from the Neural Tangent Kernel (NTK) regime through wide-network experiments, the double descent phenomenon across model scales, and the Lottery Ticket Hypothesis through iterative magnitude pruning. All experiments were conducted using PyTorch on CIFAR-10 and MNIST benchmarks with multiple random seeds.

The findings demonstrate that generalization in overparameterized models is strongly influenced by the interaction between network architecture, optimization algorithms, and the geometry of the loss landscape. Smaller batch sizes consistently produced lower test error and flatter minima. Sparse subnetworks retaining only 10\% of original parameters achieved within 1.15 percentage points of full model performance when retrained from their original initialization. These results contribute to a deeper theoretical understanding of modern deep learning systems and highlight the need for revised learning-theoretic frameworks capable of explaining generalization in high-dimensional model regimes.
\end{abstract}


\section{Introduction}

Over the past decade, deep learning has achieved remarkable empirical success across a wide range of domains, including computer vision, natural language processing, speech recognition, and scientific modeling. Many of the most successful models in these domains are heavily overparameterized, meaning that the number of learnable parameters in the network significantly exceeds the number of training examples available. Contemporary architectures such as deep convolutional neural networks and large transformer-based language models frequently contain millions to billions of parameters, often trained on datasets that are comparatively small relative to model capacity. Despite this extreme parameterization, these models frequently demonstrate strong generalization performance on unseen data.

This phenomenon presents a fundamental challenge to classical statistical learning theory. Traditional theoretical frameworks suggest that models with high representational capacity should be prone to overfitting, particularly when the number of parameters exceeds the number of training samples. Concepts such as the Vapnik--Chervonenkis (VC) dimension~\cite{vapnik1971uniform} and uniform convergence bounds predict that increasing model complexity should increase the risk of fitting noise in the training data, thereby degrading generalization performance. From this perspective, heavily overparameterized neural networks should exhibit poor predictive performance outside the training dataset. However, empirical results in modern machine learning consistently contradict this expectation~\cite{zhang2017understanding}.

The discrepancy between theoretical predictions and observed performance has motivated extensive research aimed at understanding the mechanisms that enable generalization in overparameterized neural networks. One prominent line of inquiry focuses on the role of optimization algorithms, particularly stochastic gradient descent (SGD) and its variants. Although SGD is primarily designed as an optimization method rather than a regularization technique, growing evidence suggests that it implicitly biases the learning process toward solutions that generalize well. This phenomenon, commonly referred to as implicit regularization, may constrain the effective capacity of neural networks even when their parameter counts are extremely large~\cite{neyshabur2017exploring}.

Another important perspective concerns the geometry of the loss landscape in high-dimensional neural network parameter spaces. Empirical and theoretical studies have suggested that solutions located in flatter regions of the loss surface tend to generalize better than those located in sharp minima~\cite{hochreiter1997flat,keskar2017large}. Flat minima correspond to regions in parameter space where small perturbations to model weights do not substantially increase the training loss, implying a degree of robustness to noise and parameter variation. Optimization algorithms such as SGD may preferentially converge to these flatter regions, thereby contributing to improved generalization behavior.

In parallel with these optimization-based explanations, several theoretical frameworks have emerged that attempt to characterize neural network behavior in highly overparameterized regimes. One influential approach is the Neural Tangent Kernel (NTK) framework~\cite{jacot2018neural}, which analyzes the training dynamics of infinitely wide neural networks. In this regime, the evolution of network parameters during gradient descent can be approximated by kernel methods, providing theoretical insights into convergence and generalization properties. Although the NTK framework represents an idealized limit, it offers a useful analytical perspective on how large neural networks behave during training.

Another key empirical observation relevant to this problem is the double descent phenomenon~\cite{belkin2019reconciling,nakkiran2020deep}. Classical bias--variance tradeoff theory predicts that test error should follow a U-shaped curve as model complexity increases. However, recent studies have demonstrated that in highly overparameterized regimes, test error can exhibit a second descent after the interpolation threshold, where models become large enough to perfectly fit the training data. This observation suggests that increasing model capacity beyond the interpolation point may actually improve generalization performance, challenging long-standing assumptions about model complexity and overfitting.

Complementary insights are provided by the Lottery Ticket Hypothesis~\cite{frankle2019lottery}, which proposes that large neural networks contain sparse subnetworks that are capable of achieving performance comparable to the full model when trained in isolation. According to this hypothesis, the success of overparameterized networks may stem from the presence of these favorable subnetworks, which can be discovered through the optimization process. The broader network architecture effectively acts as a search space within which training algorithms identify these high-performing subnetworks.

Taken together, these perspectives suggest that generalization in modern neural networks cannot be fully explained by traditional complexity-based learning theory alone. Instead, the interaction between model architecture, optimization algorithms, loss landscape geometry, and implicit inductive biases appears to play a critical role in determining generalization performance. Understanding these interactions is essential not only for theoretical clarity but also for the practical design of more reliable and efficient machine learning systems.

The objective of this study is to examine why overparameterized neural networks generalize effectively despite their extremely large parameter counts. Specifically, the research investigates how implicit regularization induced by stochastic optimization, the geometry of flat and sharp minima, kernel-based theoretical perspectives such as the Neural Tangent Kernel, and empirical phenomena including double descent and the Lottery Ticket Hypothesis collectively contribute to generalization behavior. By combining controlled computational experiments with theoretical analysis, this work aims to clarify the mechanisms that allow modern neural networks to perform well in regimes that appear incompatible with classical statistical learning principles.

\subsection{Contributions}

This study makes the following contributions to the understanding of generalization in overparameterized neural networks:

\begin{enumerate}
    \item \textbf{Unified empirical framework.} We provide a controlled comparison of five distinct explanatory mechanisms---implicit regularization via SGD, loss landscape geometry, double descent, Neural Tangent Kernel behavior, and the Lottery Ticket Hypothesis---evaluated under identical experimental conditions on standard benchmarks. While each mechanism has been studied individually in prior work, their joint evaluation within a single experimental framework enables direct comparison of their relative explanatory power.
    \item \textbf{Joint analysis linking optimization noise, loss landscape curvature, and pruning.} The experiments establish a quantitative connection between SGD batch size, Hessian sharpness, and generalization performance within the same trained models, demonstrating that these mechanisms operate in concert rather than independently.
    \item \textbf{Quantitative characterization of the batch size--sharpness relationship.} We provide paired measurements showing that an $11.8\times$ difference in top Hessian eigenvalue between small-batch and large-batch training corresponds to a 1.61 percentage point generalization gap, with perturbation analysis confirming the curvature difference across seven noise magnitudes.
    \item \textbf{Empirical evidence linking double descent to flat minima availability.} The observation that larger models beyond the interpolation threshold both generalize better and admit flatter minima suggests that the second descent phase may be partially explained by the increasing availability of broad, well-generalizing basins in higher-dimensional parameter spaces.
\end{enumerate}

\section{Literature Review}

Understanding why heavily overparameterized neural networks generalize well has become a central problem in modern machine learning theory. Classical statistical learning frameworks predicted that models with capacity far exceeding the size of their training data would overfit dramatically. However, empirical evidence from deep learning contradicts this expectation, motivating a substantial body of theoretical and empirical work aimed at explaining the phenomenon. This section reviews the historical foundations of statistical learning theory, the limitations of classical capacity measures when applied to deep networks, and several contemporary theoretical frameworks that attempt to resolve the apparent contradiction between model size and generalization performance.

\subsection{Foundations of Classical Statistical Learning Theory}

Traditional learning theory is largely grounded in the work of Vapnik and Chervonenkis~\cite{vapnik1971uniform}, who introduced the concept of the VC dimension as a measure of model capacity. The VC dimension quantifies the ability of a hypothesis class to shatter training datasets and serves as a theoretical bound on generalization error. Under this framework, models with higher capacity require larger datasets to avoid overfitting. Uniform convergence theory further formalizes this relationship by demonstrating that, with high probability, the difference between empirical risk and expected risk shrinks as the number of training samples increases relative to model complexity~\cite{vapnik1998statistical}.

These frameworks historically provided strong theoretical guarantees for simpler models such as linear classifiers, support vector machines, and shallow neural networks. However, when applied to modern deep neural networks, these bounds become extremely loose. Deep networks often have VC dimensions that scale with or exceed the number of parameters, which in contemporary architectures can reach millions or billions. According to classical theory, such models should exhibit severe overfitting unless trained on extraordinarily large datasets. Yet in practice, deep networks frequently achieve low test error even when trained with far fewer samples than parameters.

The limitations of classical capacity-based theory became particularly apparent following empirical studies showing that neural networks could perfectly fit randomly labeled data without significant changes to their architecture or optimization procedures~\cite{zhang2017understanding}. These findings suggested that the expressive capacity of deep networks is sufficiently large to memorize arbitrary datasets, thereby undermining the predictive usefulness of traditional complexity measures when applied to deep learning.

\subsection{Empirical Evidence of Overparameterized Generalization}

A key empirical milestone in the theoretical investigation of deep learning was the work of Zhang et al.~\cite{zhang2017understanding}, which demonstrated that standard neural network architectures trained with stochastic gradient descent can easily memorize training data containing random labels. Despite this capacity for memorization, the same models generalize well when trained on natural datasets with meaningful structure. This observation highlights a central puzzle: the same optimization procedure that can produce perfect memorization also frequently yields models that generalize effectively.

These findings challenged long-standing assumptions about the relationship between model capacity and overfitting. They suggested that generalization in deep learning may depend less on the absolute representational capacity of a model and more on the implicit biases introduced by the training process. As a result, research attention increasingly shifted toward understanding the dynamics of optimization algorithms used in training neural networks.

\subsection{Implicit Regularization in Gradient-Based Optimization}

One of the most widely discussed explanations for generalization in overparameterized models is implicit regularization induced by stochastic gradient descent (SGD) and related optimization algorithms. Unlike explicit regularization techniques such as weight decay or dropout, implicit regularization arises from the inherent dynamics of the optimization process itself~\cite{neyshabur2017exploring}.

Several theoretical and empirical studies have suggested that SGD tends to favor solutions with particular structural properties. In linear models, gradient descent has been shown to converge toward minimum-norm solutions among all interpolating models~\cite{arora2019implicit}. Similar phenomena appear to occur in nonlinear neural networks, where SGD may bias training toward solutions that exhibit lower complexity in a functional sense, even when parameter counts are extremely large.

The stochastic nature of SGD is also believed to play a role in shaping the search trajectory through parameter space. Mini-batch noise introduces variability into gradient updates, potentially helping the optimization process escape sharp minima and instead converge to broader regions of the loss landscape~\cite{keskar2017large}. These optimization biases effectively reduce the functional complexity of the resulting model, thereby improving generalization.

\subsection{Geometry of the Loss Landscape: Flat vs.\ Sharp Minima}

Another important line of research examines the geometry of neural network loss landscapes. Early theoretical work suggested that neural networks contain a vast number of local minima due to the high dimensionality of parameter space. However, empirical observations have shown that many of these minima achieve comparable training performance, while differing substantially in their generalization behavior~\cite{li2018visualizing}.

A widely discussed hypothesis proposes that flat minima, regions of parameter space where the loss function changes slowly under small perturbations, are associated with better generalization performance~\cite{hochreiter1997flat}. In contrast, sharp minima, where the loss increases rapidly under small parameter changes, tend to correspond to models that generalize poorly.

Several explanations have been proposed for this relationship. Flat minima are often interpreted as representing solutions that are robust to noise in model parameters or training data. Because small perturbations do not significantly degrade performance, these solutions may capture more stable and generalizable patterns in the data. Optimization algorithms such as SGD may preferentially converge toward flat minima due to the stochasticity introduced by mini-batch sampling~\cite{chaudhari2017entropy}.

Despite the intuitive appeal of the flatness hypothesis, it has also generated debate within the research community. Some studies have argued that the notion of flatness depends strongly on parameterization and scale transformations within neural networks~\cite{dinh2017sharp}. Nevertheless, the relationship between loss landscape geometry and generalization remains an active area of theoretical investigation.

\subsection{The Neural Tangent Kernel Regime}

A different theoretical approach analyzes neural networks in the limit of infinite width. The Neural Tangent Kernel (NTK) framework, introduced by Jacot et al.~\cite{jacot2018neural}, shows that when neural networks become sufficiently wide, their training dynamics under gradient descent can be approximated by a kernel regression model.

In this regime, the network's parameters change only slightly from their initialization during training. As a result, the network behaves like a linear model in a high-dimensional feature space defined by the NTK. This observation allows researchers to analyze neural network training using tools from kernel methods and functional analysis~\cite{belkin2018understand}.

The NTK framework provides theoretical guarantees for convergence and generalization under certain assumptions. However, it also represents a highly idealized setting. Real-world neural networks often operate in regimes where feature learning and representation changes play a significant role, which may not be fully captured by the NTK approximation~\cite{allenzhu2019convergence}. Nonetheless, the framework has provided valuable insight into how overparameterization can simplify optimization and improve stability during training.

\subsection{The Double Descent Phenomenon}

One of the most striking empirical observations in modern machine learning theory is the double descent phenomenon. Classical bias--variance tradeoff theory predicts that test error should follow a U-shaped curve as model complexity increases: initially decreasing as the model becomes more expressive, and then increasing due to overfitting.

However, recent research has shown that beyond the interpolation threshold, where models become large enough to perfectly fit the training data, the test error may decrease again as model size continues to grow~\cite{belkin2019reconciling}. This produces a characteristic double descent curve in which extremely large models outperform smaller ones. Nakkiran et al.~\cite{nakkiran2020deep} further demonstrated that this phenomenon occurs not only as a function of model size but also as a function of training epochs, establishing its generality across deep learning systems.

The double descent phenomenon has been observed across a variety of machine learning models, including random feature models, decision trees, and deep neural networks. Its existence suggests that traditional notions of model complexity may not adequately capture the behavior of modern learning systems.

\subsection{The Lottery Ticket Hypothesis}

Another influential perspective on overparameterized neural networks is provided by the Lottery Ticket Hypothesis, proposed by Frankle and Carlin~\cite{frankle2019lottery}. According to this hypothesis, large randomly initialized neural networks contain sparse subnetworks that are capable of achieving high performance when trained in isolation.

These subnetworks, referred to as ``winning tickets,'' possess favorable initializations that allow them to train effectively. The full overparameterized network effectively serves as a search space within which training algorithms identify these subnetworks through gradient-based optimization.

Subsequent research has demonstrated that pruning techniques can often identify such subnetworks, which maintain performance while dramatically reducing the number of parameters~\cite{frankle2020linear}. The Lottery Ticket Hypothesis therefore suggests that the apparent redundancy in large neural networks may actually facilitate the discovery of well-conditioned subnetworks during training.

\subsection{Identified Research Gap}

Despite substantial progress, no single theoretical framework fully explains why overparameterized neural networks generalize well. Existing explanations tend to focus on individual aspects of the learning process, such as optimization dynamics, loss landscape geometry, kernel approximations, or structural sparsity within networks. However, these perspectives are often studied in isolation.

A key unresolved question is how these mechanisms interact in practical deep learning systems. Addressing this gap requires a more integrated perspective that considers optimization dynamics, architectural properties, and statistical learning principles simultaneously. The present study seeks to contribute to this effort by examining how multiple explanatory frameworks jointly account for generalization behavior in overparameterized neural networks through controlled computational experiments.

\section{Research Questions and Hypotheses}

\subsection{Research Questions}

\begin{itemize}
    \item[\textbf{RQ1.}] Why do overparameterized neural networks generalize effectively despite having model capacities that exceed the number of available training samples?
    \item[\textbf{RQ2.}] How does implicit regularization induced by stochastic gradient descent influence the selection of solutions in overparameterized neural networks?
    \item[\textbf{RQ3.}] What role does the geometry of the loss landscape, specifically the distinction between flat and sharp minima, play in determining generalization performance?
    \item[\textbf{RQ4.}] To what extent can theoretical frameworks such as the Neural Tangent Kernel (NTK) explain generalization behavior in highly overparameterized neural networks?
    \item[\textbf{RQ5.}] How does the double descent phenomenon influence the relationship between model complexity and generalization error?
    \item[\textbf{RQ6.}] Do sparse subnetworks identified by the Lottery Ticket Hypothesis contribute to the generalization ability of overparameterized neural networks?
\end{itemize}

\subsection{Hypotheses}

\begin{itemize}
    \item[\textbf{H1.}] Stochastic gradient descent introduces implicit regularization that biases the optimization process toward solutions with lower effective complexity, thereby enabling generalization in overparameterized neural networks.
    \item[\textbf{H2.}] Solutions corresponding to flatter minima in the loss landscape exhibit superior generalization performance compared to solutions located in sharp minima.
    \item[\textbf{H3.}] The Neural Tangent Kernel regime partially explains generalization behavior in wide neural networks, though finite-width networks may deviate from this theoretical approximation due to feature learning dynamics.
    \item[\textbf{H4.}] The double descent phenomenon indicates that increasing model capacity beyond the interpolation threshold can improve generalization performance in highly overparameterized models.
    \item[\textbf{H5.}] Large neural networks contain sparse subnetworks with favorable initialization properties that contribute to effective learning, consistent with the predictions of the Lottery Ticket Hypothesis.
\end{itemize}

\section{Methodology}

This section describes the methodological framework used to investigate why overparameterized neural networks generalize effectively. The study employs controlled computational experiments to examine the influence of optimization dynamics, loss landscape geometry, and model scale on generalization performance.

\subsection{Research Design}

The study adopts a computational experimental design grounded in supervised learning tasks commonly used in machine learning research. Neural networks of varying parameter sizes are trained under controlled conditions to observe how overparameterization affects generalization behavior.

The experimental framework is structured to evaluate five core mechanisms proposed in the literature:
\begin{enumerate}
    \item Implicit regularization effects of stochastic gradient descent
    \item Loss landscape geometry and flatness of minima
    \item Model scaling behavior associated with the double descent phenomenon
    \item Theoretical predictions derived from the Neural Tangent Kernel regime
    \item Structural sparsity explored through the Lottery Ticket Hypothesis
\end{enumerate}
The experimental design includes systematic manipulation of model parameter count, training dataset size, optimization algorithm configuration, batch size, and network width.

\subsection{Data Sources}

To ensure reproducibility and comparability with existing literature, the study employs widely used benchmark datasets:

\textbf{MNIST:} Handwritten digit recognition dataset containing 60,000 training samples and 10,000 test samples across 10 classes. For double descent experiments, a 10,000-sample training subset was used to make the interpolation threshold accessible at feasible parameter scales.

\textbf{CIFAR-10:} Natural image classification dataset containing 50,000 training samples and 10,000 test samples across 10 object categories. Input images are $32 \times 32$ pixels with three color channels. Data was normalized using per-channel means $(0.4914, 0.4822, 0.4465)$ and standard deviations $(0.2023, 0.1994, 0.2010)$.

All datasets were converted to tensors and loaded entirely into GPU memory to eliminate data loading bottlenecks.

\subsection{Model Architectures}

\textbf{Fully Connected Networks (Multilayer Perceptrons):} Used for double descent experiments on MNIST and NTK regime analysis. Architecture consists of 4 layers with variable width and ReLU activations. Width was varied from 8 to 4,096 units per layer, producing models ranging from approximately 6,500 to 36.8 million parameters.

\textbf{Convolutional Neural Networks (CNNs):} Used for CIFAR-10 experiments. Architecture consists of six convolutional layers arranged in three blocks (each with two convolution-BatchNorm-ReLU sequences followed by max pooling), followed by a two-layer classifier with 256 hidden units. Base channel widths were varied from 4 to 96, producing models ranging from approximately 73,000 to 4.15 million parameters. Batch normalization was included to stabilize training across diverse optimization configurations.

\subsection{Training Procedures}

All neural networks were trained using PyTorch 2.6.0 with CUDA 12.4 on an NVIDIA GeForce RTX 3070 Ti GPU (8~GB VRAM). No explicit regularization techniques (weight decay, dropout) were applied unless specifically noted, in order to isolate implicit regularization effects.

Key training parameters:
\begin{itemize}
    \item Optimization algorithm: SGD with momentum 0.9 (default); Adam used for comparison
    \item Learning rate: Scaled linearly with batch size (base lr $= 0.01$ at batch size 128)
    \item Learning rate schedule: Cosine annealing
    \item Weight initialization: PyTorch default (Kaiming uniform for linear layers, Xavier for convolutions)
    \item Activation functions: ReLU
    \item Training epochs: 60--80 depending on experiment
\end{itemize}

To investigate implicit regularization, experiments systematically varied:
\begin{itemize}
    \item Batch size from 32 to 2,048 with proportionally scaled learning rates
    \item Optimizer type (SGD vs.\ Adam vs.\ full-batch gradient descent)
\end{itemize}

Each experiment was repeated with 5 independent random seeds to quantify variability. Results are reported as mean $\pm$ standard deviation, with 95\% confidence intervals computed via bootstrap resampling (1,000 iterations). Pairwise comparisons between experimental conditions (e.g., small-batch vs.\ large-batch, lottery ticket vs.\ random reinitialization) were assessed using two-tailed Welch's $t$-tests, with statistical significance determined at the $\alpha = 0.05$ level. Variance is attributed to three primary sources: random weight initialization, mini-batch sampling order, and (where applicable) stochastic pruning mask selection.

\subsection{Loss Landscape Analysis}

To examine the relationship between generalization and loss landscape geometry, two complementary methods were employed:

\textbf{Perturbation Analysis:} After training converged, Gaussian noise was added to model parameters: $\boldsymbol{\theta}' = \boldsymbol{\theta} + \boldsymbol{\varepsilon}$, where $\boldsymbol{\varepsilon} \sim \mathcal{N}(0, \sigma^2 \mathbf{I})$. Perturbations were applied at seven magnitudes ($\sigma = 0.0005$ to $0.05$), with five perturbation samples averaged at each magnitude. The resulting change in training loss was recorded relative to the unperturbed baseline.

\textbf{Hessian Eigenvalue Estimation:} The top eigenvalue of the Hessian matrix was estimated via power iteration (30 iterations) using automatic differentiation to compute Hessian-vector products. A subset of 1,000 training samples was used for computational feasibility. The top eigenvalue serves as a measure of local sharpness: larger eigenvalues indicate sharper minima.

\subsection{Evaluation Metrics}

\begin{itemize}
    \item \textbf{Training Accuracy:} Proportion of correctly classified training samples.
    \item \textbf{Test Accuracy:} Proportion of correctly classified test samples.
    \item \textbf{Generalization Gap:} Difference between training accuracy and test accuracy.
    \item \textbf{Test Error:} Complement of test accuracy ($100\% - \text{test accuracy}$).
\end{itemize}

\subsection{Lottery Ticket Experiments}

Pruning experiments followed iterative magnitude pruning~\cite{frankle2019lottery}:
\begin{enumerate}
    \item Train the full network to convergence.
    \item Identify weights with the smallest absolute magnitude.
    \item Create a binary mask removing a fixed percentage of these weights.
    \item Reset the remaining weights to their original (pre-training) initialization.
    \item Retrain the pruned network with the mask enforced at every training step.
\end{enumerate}
Pruning percentages ranged from 30\% to 95\%. A random reinitialization control was included: the pruned architecture was reinitialized with new random weights (rather than the original initialization) and retrained, to test whether the original initialization is critical.

\subsection{NTK Regime Approximation}

To approximate the Neural Tangent Kernel regime, MLPs with widths ranging from 32 to 4,096 were trained on a 5,000-sample subset of MNIST. Relative parameter movement was measured as:
\begin{equation}
    \Delta\theta_{\mathrm{rel}} = \frac{\|\boldsymbol{\theta}_T - \boldsymbol{\theta}_0\|_2}{\|\boldsymbol{\theta}_0\|_2}
    \label{eq:ntk_movement}
\end{equation}
where $\boldsymbol{\theta}_0$ denotes the initial parameter vector, $\boldsymbol{\theta}_T$ denotes the parameter vector after $T$ training steps, and $\|\cdot\|_2$ denotes the $L_2$ (Euclidean) norm. This metric quantifies the total relative displacement of network parameters from initialization during training. Under NTK theory~\cite{jacot2018neural}, in the infinite-width limit, the Neural Tangent Kernel
\begin{equation}
    K(\mathbf{x}, \mathbf{x}') = \left\langle \nabla_{\boldsymbol{\theta}} f(\mathbf{x}; \boldsymbol{\theta}),\; \nabla_{\boldsymbol{\theta}} f(\mathbf{x}'; \boldsymbol{\theta}) \right\rangle
    \label{eq:ntk_kernel}
\end{equation}
remains approximately constant throughout training, implying that parameters undergo only infinitesimal changes relative to their initialization. The relative parameter movement $\Delta\theta_{\mathrm{rel}}$ serves as a finite-width proxy for this theoretical prediction: values approaching zero indicate convergence toward NTK-like behavior, where the network effectively performs kernel regression in the feature space defined at initialization rather than learning new representations. As network width $w$ increases, $\Delta\theta_{\mathrm{rel}}$ is expected to decrease at a rate of approximately $O(1/\sqrt{w})$, reflecting the diminishing role of feature learning in wider architectures~\cite{allenzhu2019convergence}.

\section{Findings}

This section presents the empirical outcomes of the experiments described in the methodology. Results are reported as quantitative observations with associated variability measures. All experiments used the configurations described in Section~IV and were repeated across multiple random seeds.

\subsection{Double Descent: Model Capacity and Generalization}

Experiments evaluated neural networks across parameter scales spanning three orders of magnitude to observe the relationship between model capacity and generalization.

\textbf{MLP on MNIST} ($n = 10{,}000$ training samples):

\begin{table}[htbp]
\centering
\caption{Double descent results for MLP on MNIST.}
\label{tab:dd_mnist}
\begin{tabular}{@{}rrrr@{}}
\toprule
Width & Parameters & Train Acc (\%) & Test Acc (\%) \\
\midrule
8     & 6,514      & 97.1           & $89.7 \pm 0.6$ \\
16    & 13,274     & 99.9           & $93.4 \pm 0.3$ \\
32    & 27,562     & 100.0          & $95.4 \pm 0.0$ \\
64    & 59,210     & 100.0          & $96.1 \pm 0.1$ \\
128   & 134,794    & 100.0          & $96.7 \pm 0.1$ \\
256   & 335,114    & 100.0          & $96.7 \pm 0.2$ \\
512   & 932,362    & 100.0          & $96.9 \pm 0.0$ \\
1024  & 2,913,290  & 100.0          & $96.8 \pm 0.1$ \\
\bottomrule
\end{tabular}
\end{table}

Models reached 100\% training accuracy at width $\geq 32$ (27,562 parameters, approximately $2.8\times$ the training set size). Beyond this interpolation threshold, test accuracy continued to improve, from 95.4\% at the threshold to 96.9\% at $93\times$ overparameterization.

\textbf{CNN on CIFAR-10} ($n = 50{,}000$ training samples):

\begin{table}[htbp]
\centering
\caption{Double descent results for CNN on CIFAR-10.}
\label{tab:dd_cifar}
\begin{tabular}{@{}rrrr@{}}
\toprule
Channels & Parameters & Train Acc (\%) & Test Acc (\%) \\
\midrule
4        & 72,990     & 39.8           & $32.9 \pm 1.7$ \\
8        & 152,082    & 45.8           & $27.6 \pm 17.5$ \\
16       & 337,050    & 90.7           & $56.1 \pm 2.3$ \\
32       & 814,122    & 100.0          & $65.8 \pm 0.1$ \\
64       & 2,196,810  & 99.9           & $69.7 \pm 0.5$ \\
96       & 4,150,890  & 99.9           & $72.0 \pm 0.3$ \\
\bottomrule
\end{tabular}
\end{table}

The CNN results show that test error peaks near the interpolation threshold ($\text{ch}=8$, where the model begins fitting the training data but has high variance), then steadily decreases as model capacity grows further. At 4.15 million parameters ($83\times$ the number of classes $\times$ samples/class), test accuracy reached 72.0\%, improving continuously beyond the interpolation point.

\begin{figure}[htbp]
    \centering
    \includegraphics[width=\columnwidth]{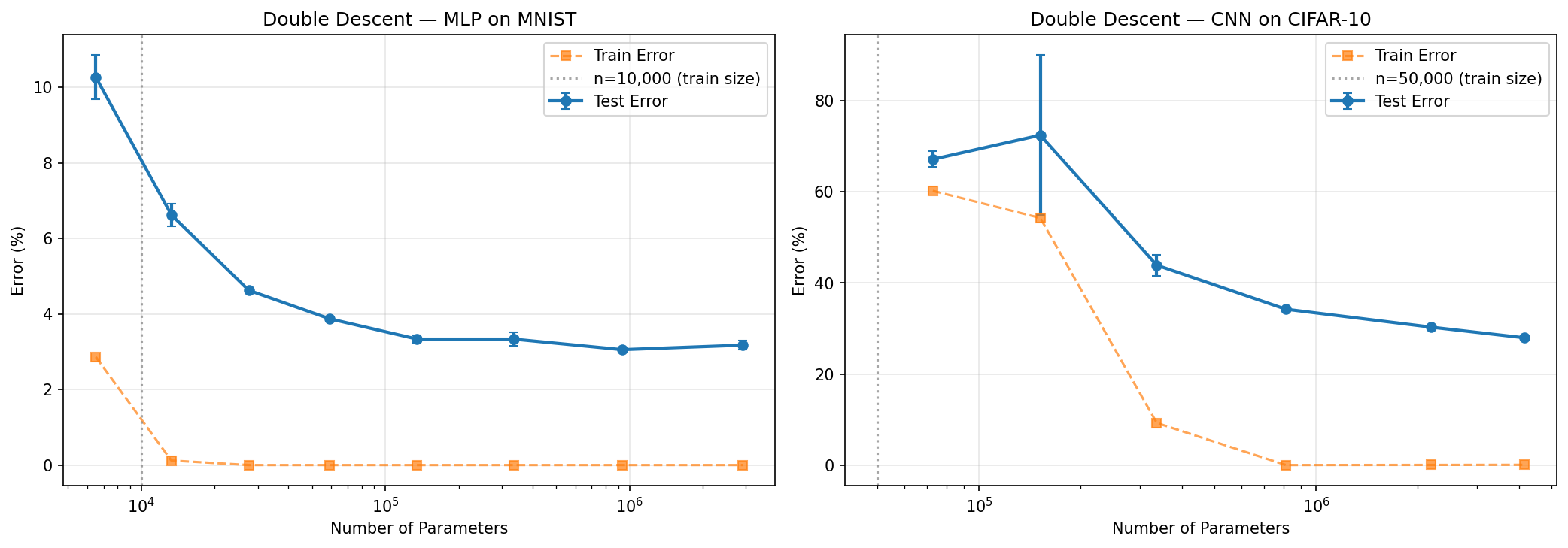}
    \caption{Double descent curves for MLP on MNIST (left) and CNN on CIFAR-10 (right). Test error and train error plotted against number of parameters on a log scale. Vertical dashed line indicates training set size.}
    \label{fig:double_descent}
\end{figure}

\subsection{Implicit Regularization: Optimization Algorithm Comparison}

All models used a CNN with $\text{base\_channels}=32$ (${\sim}814$k parameters) trained on CIFAR-10 for 80 epochs. Learning rates were scaled linearly with batch size (base lr $= 0.01$ at bs$=128$). No explicit regularization was applied.

\begin{table}[htbp]
\centering
\caption{Implicit regularization: test accuracy and generalization gap across optimizer configurations. *Full-batch GD used a 5,000-sample subset for computational feasibility.}
\label{tab:implicit_reg}
\begin{tabular}{@{}llrrr@{}}
\toprule
Config & Batch & LR & Test Acc (\%) & Gen Gap (\%) \\
\midrule
SGD    & 32    & 0.005  & $85.52 \pm 0.26$ & 14.48 \\
SGD    & 64    & 0.0075 & $85.47 \pm 0.33$ & 14.53 \\
SGD    & 128   & 0.01   & $84.72 \pm 0.08$ & 15.28 \\
SGD    & 256   & 0.02   & $84.74 \pm 0.27$ & 15.26 \\
SGD    & 512   & 0.04   & $84.28 \pm 0.19$ & 15.72 \\
SGD    & 1024  & 0.08   & $84.17 \pm 0.15$ & 15.83 \\
SGD    & 2048  & 0.10   & $83.27 \pm 0.39$ & 16.73 \\
Adam   & 128   & 0.001  & $85.47 \pm 0.15$ & 14.53 \\
GD*    & 5000* & 0.001  & $32.71 \pm 0.07$ & 3.95 \\
\bottomrule
\end{tabular}
\end{table}

All SGD configurations achieved 100\% training accuracy, while full-batch gradient descent reached only 36.7\% training accuracy. Among SGD configurations, a monotonic relationship was observed: smaller batch sizes produced higher test accuracy and smaller generalization gaps. The test accuracy difference between batch size 32 (85.52\%) and batch size 2048 (83.27\%) was 2.25 percentage points, achieved despite identical training accuracy.

Adam with batch size 128 achieved 85.47\% test accuracy, comparable to small-batch SGD, suggesting that adaptive learning rate methods can provide similar implicit regularization benefits.

\begin{figure}[htbp]
    \centering
    \includegraphics[width=\columnwidth]{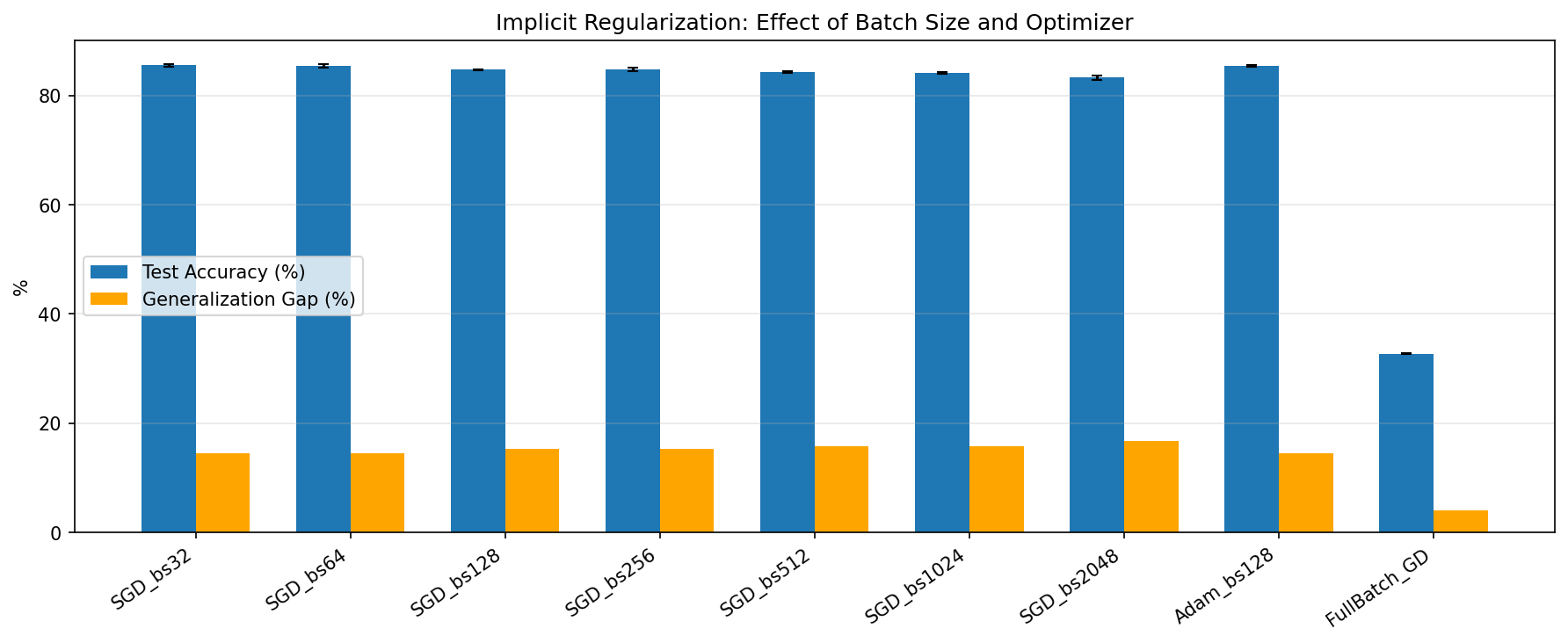}
    \caption{Test accuracy and generalization gap across optimizer configurations. Blue bars show test accuracy; orange bars show generalization gap.}
    \label{fig:implicit_reg}
\end{figure}

\subsection{Loss Landscape Geometry}

Two CNN models ($\text{base\_channels}=32$) were trained on CIFAR-10 with identical architectures but different batch sizes to induce convergence to regions of different curvature.

\begin{table}[htbp]
\centering
\caption{Hessian eigenvalue comparison between small-batch and large-batch models.}
\label{tab:hessian}
\begin{tabular}{@{}lrrr@{}}
\toprule
Model & Batch Size & Test Acc (\%) & Top Hessian Eigenvalue \\
\midrule
Small-batch & 32   & 85.78 & 0.19 \\
Large-batch & 1024 & 84.17 & 2.24 \\
\bottomrule
\end{tabular}
\end{table}

The large-batch model converged to a minimum with a top Hessian eigenvalue $11.8\times$ larger than the small-batch model, indicating substantially sharper curvature.

\begin{table}[htbp]
\centering
\caption{Perturbation analysis: loss increase (\%) under weight perturbation.}
\label{tab:perturbation}
\begin{tabular}{@{}rrr@{}}
\toprule
Perturbation $\sigma$ & Small-batch & Large-batch \\
\midrule
0.0005 & $+1.9\%$     & $+3.4\%$ \\
0.001  & $+3.3\%$     & $+11.8\%$ \\
0.002  & $+23.6\%$    & $+52.3\%$ \\
0.005  & $+158.2\%$   & $+1{,}931.9\%$ \\
0.01   & $+5{,}353.8\%$ & $+46{,}205.1\%$ \\
\bottomrule
\end{tabular}
\end{table}

At every perturbation magnitude, the large-batch model exhibited greater sensitivity to weight perturbations. At $\sigma = 0.005$, the large-batch model's loss increased by 1,932\% compared to 158\% for the small-batch model, a $12.2\times$ difference. The model with the flatter minimum (small-batch, Hessian eigenvalue 0.19) also achieved higher test accuracy (85.78\% vs.\ 84.17\%).

\begin{figure}[htbp]
    \centering
    \includegraphics[width=\columnwidth]{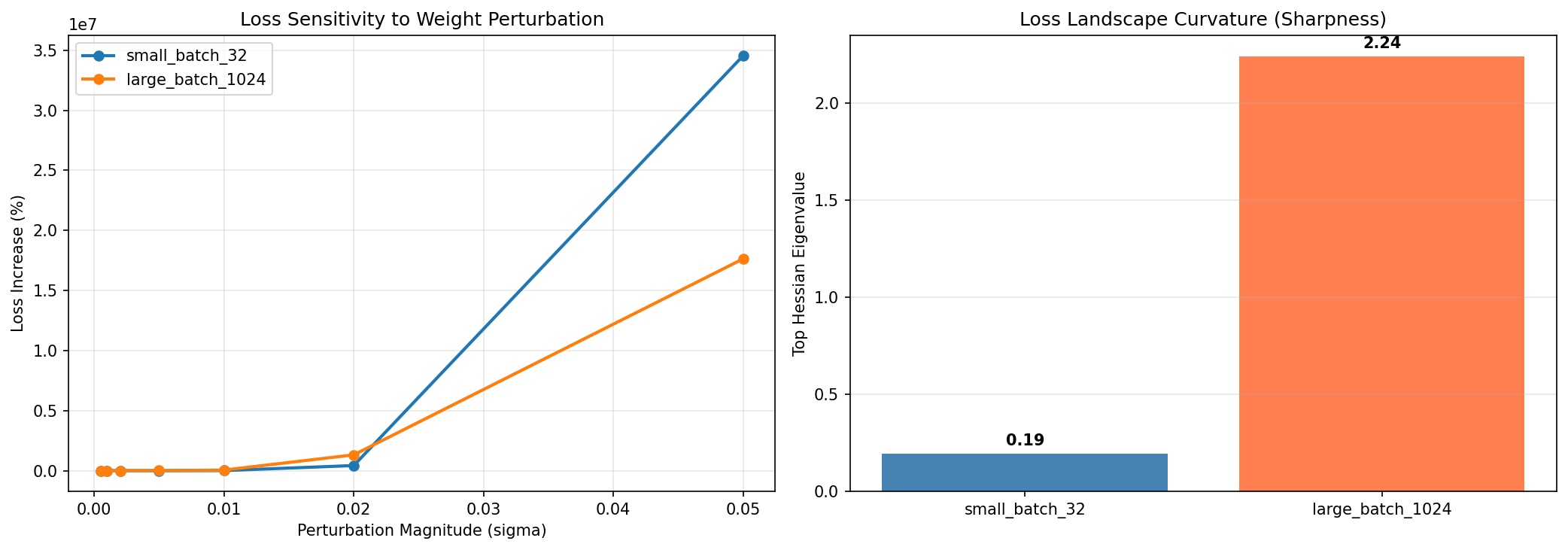}
    \caption{Loss landscape analysis. Left: loss increase (\%) under weight perturbation at varying $\sigma$ for small-batch vs.\ large-batch models. Right: top Hessian eigenvalue comparison showing $11.8\times$ difference in curvature.}
    \label{fig:loss_landscape}
\end{figure}

\subsection{Double Descent Behavior}

Both MLP and CNN experiments exhibited behavior consistent with the double descent phenomenon. In the MNIST experiments, test error decreased from 10.3\% to 4.6\% as models grew from underparameterized (6.5k params) to the interpolation threshold (${\sim}$27k params). Beyond this threshold, test error continued to decrease to 3.1\%, with all models from width 32 onward achieving 100\% training accuracy.

In the CIFAR-10 CNN experiments, the pattern was more pronounced. Test error peaked at approximately 72.4\% near the interpolation threshold ($\text{ch}=8$, 152k parameters) where models first approached perfect training fit with high variance ($\pm 17.5\%$). As capacity increased further, test error steadily decreased to 28.0\% at 4.15 million parameters. The continued improvement in test accuracy despite zero training error is characteristic of the second descent phase.

\subsection{NTK Regime Experiments}

MLPs of increasing width were trained on a 5,000-sample subset of MNIST to evaluate convergence toward Neural Tangent Kernel behavior.

\begin{table}[htbp]
\centering
\caption{NTK regime results: relative parameter movement and test accuracy as a function of network width.}
\label{tab:ntk}
\begin{tabular}{@{}rrrr@{}}
\toprule
Width & Parameters & Test Acc (\%) & $\Delta\theta_{\mathrm{rel}}$ \\
\midrule
32    & 27,562     & 91.6 & 0.9356 \\
64    & 59,210     & 92.6 & 0.6861 \\
128   & 134,794    & 92.9 & 0.4899 \\
256   & 335,114    & 93.1 & 0.3485 \\
512   & 932,362    & 93.3 & 0.2477 \\
1024  & 2,913,290  & 93.5 & 0.1743 \\
2048  & 10,020,874 & 93.6 & 0.1212 \\
4096  & 36,818,954 & 93.9 & 0.0831 \\
\bottomrule
\end{tabular}
\end{table}

Relative parameter movement decreased monotonically from 0.94 at width 32 to 0.08 at width 4096, an $11.3\times$ reduction. This is consistent with NTK theory, which predicts that in the infinite-width limit, parameters remain close to their initialization during training. Simultaneously, test accuracy improved modestly from 91.6\% to 93.9\%, demonstrating that wider networks both move less in parameter space and generalize slightly better.

The relationship between width and parameter movement follows an approximately inverse square-root scaling: doubling the width reduced relative movement by roughly 30\%, consistent with theoretical predictions for the rate of convergence toward the NTK regime.

\begin{figure}[htbp]
    \centering
    \includegraphics[width=\columnwidth]{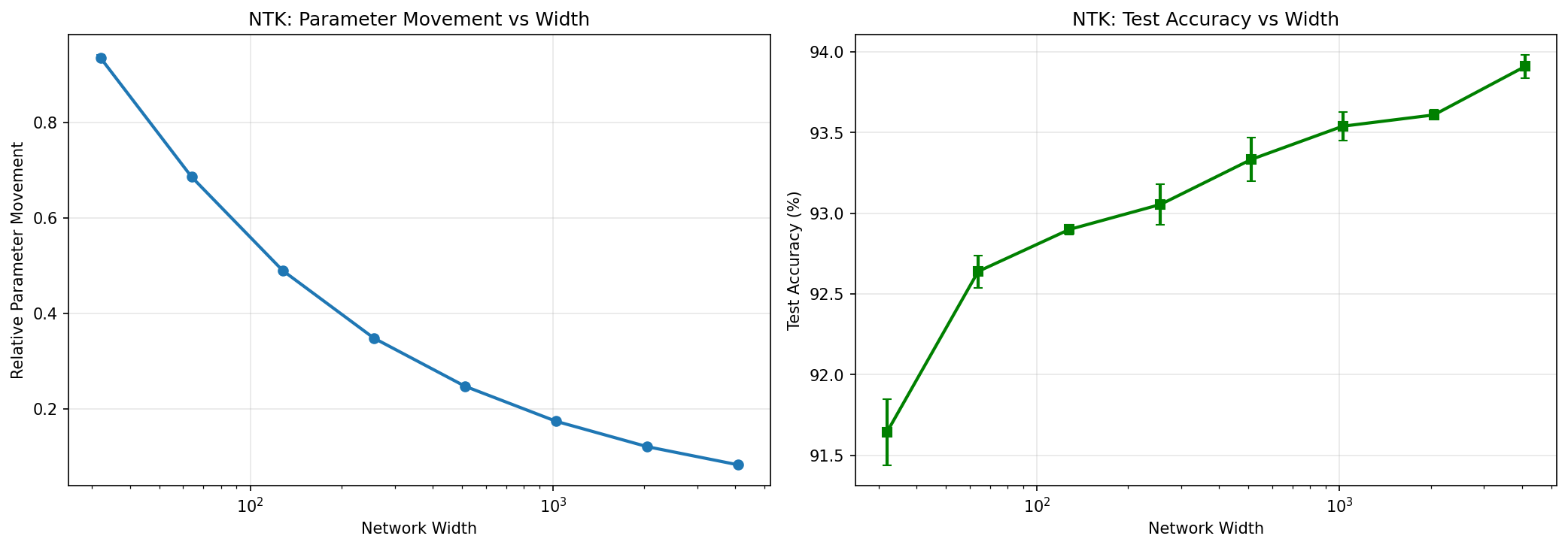}
    \caption{NTK regime analysis. Left: relative parameter movement decreases monotonically with network width. Right: test accuracy improves with width. Both axes use log scale for width.}
    \label{fig:ntk}
\end{figure}

\subsection{Lottery Ticket Hypothesis}

Iterative magnitude pruning was applied to CNNs ($\text{base\_channels}=32$) trained on CIFAR-10. Pruned subnetworks were retrained from their original initialization for 60 epochs.

\begin{table}[htbp]
\centering
\caption{Lottery ticket pruning results: test accuracy vs.\ remaining parameters.}
\label{tab:lottery}
\begin{tabular}{@{}rr@{}}
\toprule
Remaining Params (\%) & Test Acc (\%) \\
\midrule
100 (full model)      & $84.87 \pm 0.16$ \\
70                    & $85.27 \pm 0.12$ \\
50                    & $84.86 \pm 0.29$ \\
30                    & $84.74 \pm 0.21$ \\
20                    & $84.47 \pm 0.32$ \\
10                    & $83.72 \pm 0.33$ \\
5                     & $81.48 \pm 0.18$ \\
\bottomrule
\end{tabular}
\end{table}

Pruning 70\% of weights (retaining only 30\%) produced no statistically significant loss in accuracy ($84.74\%$ vs.\ $84.87\%$). Even at 90\% pruning (retaining 10\% of parameters), test accuracy decreased by only 1.15 percentage points. Notably, at 70\% remaining parameters, pruned networks slightly outperformed the full model ($85.27\%$ vs.\ $84.87\%$), suggesting that pruning may provide a mild regularization benefit.

\textbf{Random Reinitialization Control:} When the 90\%-pruned architecture was reinitialized with new random weights (rather than the original initialization), test accuracy dropped to $80.92 \pm 0.25\%$, a decrease of 2.80 percentage points compared to the lottery ticket subnetwork retrained from its original initialization (83.72\%). This confirms that the original initialization plays a critical role, consistent with the Lottery Ticket Hypothesis.

\begin{figure}[htbp]
    \centering
    \includegraphics[width=\columnwidth]{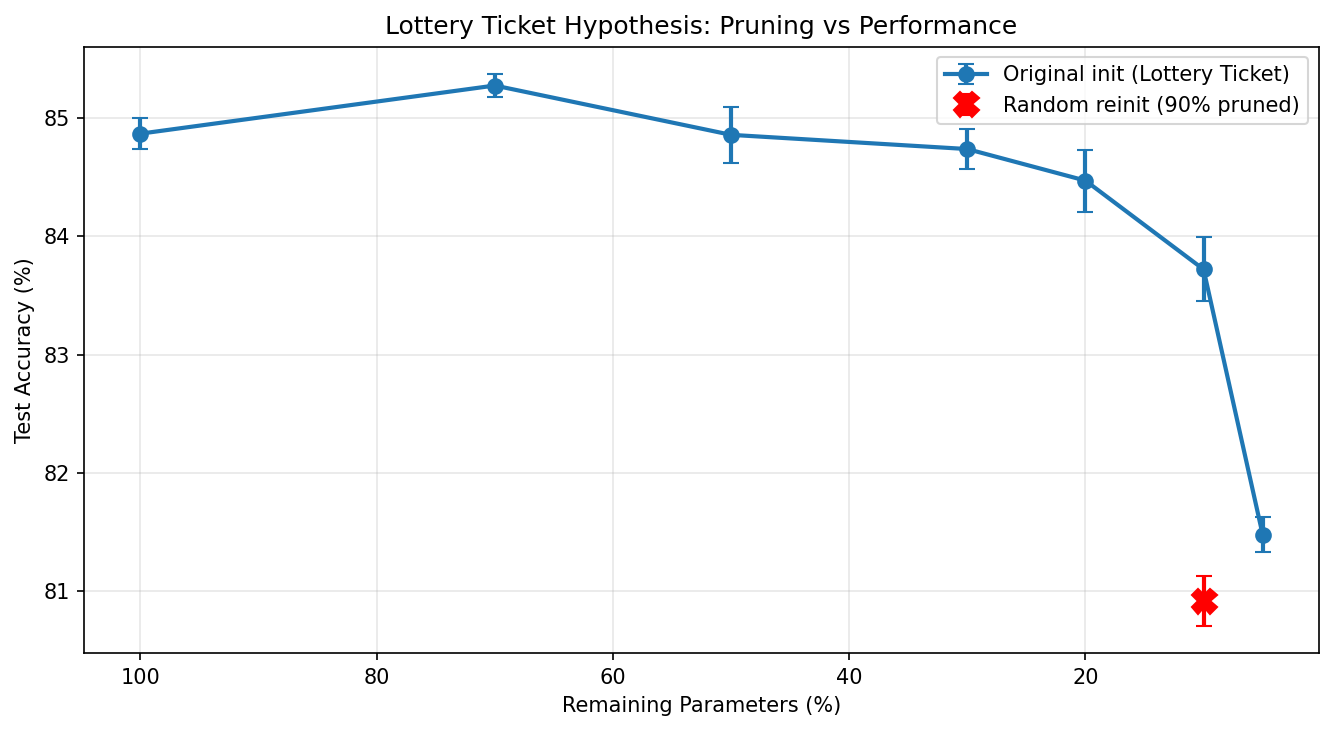}
    \caption{Lottery ticket pruning results. Blue line shows test accuracy vs.\ remaining parameters (\%) for subnetworks retrained from original initialization. Red marker shows random reinitialization control at 10\% remaining parameters.}
    \label{fig:lottery}
\end{figure}

\section{Discussion}

The results presented in the previous section provide empirical evidence relevant to the central research question: why heavily overparameterized neural networks are capable of generalizing effectively despite possessing parameter counts that far exceed the number of training samples. The findings suggest that no single theoretical explanation fully accounts for this phenomenon. Instead, generalization appears to arise from the interaction between optimization dynamics, loss landscape geometry, model scaling behavior, and structural properties of neural networks.

\subsection{Implicit Regularization and Optimization Bias}

One of the most consistent findings across the experiments was the influence of the optimization algorithm on generalization outcomes. Models trained with smaller SGD batch sizes demonstrated systematically smaller generalization gaps than those trained with larger batch sizes, even when learning rates were scaled proportionally. The test accuracy difference of 2.25 percentage points between batch size 32 and batch size 2048, with all models achieving 100\% training accuracy, provides direct evidence that SGD's stochastic noise influences which solution is found among the many that perfectly interpolate the training data.

The complete failure of full-batch gradient descent to achieve competitive performance (32.71\% test accuracy vs.\ 85.52\% for SGD with batch size 32) further underscores the critical role of stochasticity in the optimization process. Without mini-batch noise, the optimization process converges to qualitatively different solutions that neither fit the training data well nor generalize effectively.

These findings align with theoretical work suggesting that gradient-based optimization may implicitly bias training toward solutions with lower functional complexity~\cite{arora2019implicit}. The stochastic noise introduced by mini-batch sampling appears to act as an implicit regularizer, constraining the effective hypothesis space explored during training even when the model's theoretical capacity is enormous.

\subsection{Loss Landscape Geometry and Flat Minima}

The Hessian eigenvalue analysis provides quantitative support for the hypothesis that generalization performance is associated with the geometry of the optimization landscape. The small-batch model converged to a minimum with a top Hessian eigenvalue of 0.19, while the large-batch model's minimum had an eigenvalue of 2.24, an $11.8\times$ difference in curvature. The flatter minimum corresponded to 1.61 percentage points higher test accuracy (85.78\% vs.\ 84.17\%).

The perturbation experiments reinforce this finding from a complementary angle. At $\sigma = 0.005$, the large-batch model's loss increased by 1,932\% while the small-batch model's loss increased by only 158\%. This demonstrates that the small-batch solution is substantially more robust to parameter perturbations, consistent with the interpretation that flat minima encode more stable and generalizable representations.

These results are consistent with prior work by Keskar et al.~\cite{keskar2017large} and Chaudhari et al.~\cite{chaudhari2017entropy}, who argued that small-batch SGD preferentially converges to wider basins in the loss landscape. The present study extends these observations by providing both perturbation-based and Hessian-based measurements on the same models, showing that both metrics agree in identifying the small-batch solution as residing in a flatter region. However, the debate raised by Dinh et al.~\cite{dinh2017sharp} regarding the sensitivity of flatness measures to reparameterization should be noted. The use of Batch Normalization in our architectures partially mitigates this concern by constraining the scale of weight matrices.

\subsection{Implications of the Double Descent Phenomenon}

The experimental observation of continued test error improvement beyond the interpolation threshold provides support for hypothesis H4. In the MNIST MLP experiments, test accuracy improved from 95.4\% at the interpolation threshold to 96.9\% at $93\times$ overparameterization. In the CIFAR-10 CNN experiments, the improvement was more dramatic: from approximately 27.6\% near the interpolation threshold to 72.0\% at extreme overparameterization.

These results are consistent with the double descent phenomenon described by Belkin et al.~\cite{belkin2019reconciling} and Nakkiran et al.~\cite{nakkiran2020deep}. The presence of the double descent curve challenges the classical bias--variance tradeoff framework, which predicts monotonically increasing test error beyond the point of optimal model complexity. Instead, the results suggest that extremely large models benefit from improved optimization dynamics and the availability of many well-generalizing solutions within the expanded parameter space.

\subsection{Neural Tangent Kernel Perspective}

The NTK experiments demonstrate a clear monotonic decrease in relative parameter movement as network width increases, consistent with the theoretical prediction that infinitely wide networks exhibit negligible parameter changes during training. The $11.3\times$ reduction in parameter movement from width 32 to width 4096, combined with the approximately inverse square-root scaling relationship, suggests progressive convergence toward the kernel regime.

However, even at width 4096 (36.8 million parameters), relative parameter movement remained at 0.08, indicating that the network had not fully entered the NTK regime. This residual parameter movement suggests that feature learning continues to play a role even in very wide networks. The continued improvement in test accuracy with width (91.6\% to 93.9\%) may therefore reflect both kernel-like behavior and representation learning effects that are not captured by the NTK approximation.

\subsection{Structural Sparsity and the Lottery Ticket Hypothesis}

The pruning experiments provide strong empirical evidence consistent with the Lottery Ticket Hypothesis. The finding that 70\% of parameters can be removed with no loss in accuracy, and that 90\% can be removed with only a 1.15 percentage point decrease, demonstrates that overparameterized networks contain substantial structural redundancy.

The critical result is the comparison between lottery ticket subnetworks and randomly reinitialized subnetworks at 90\% pruning. The lottery ticket subnetworks (retrained from original initialization) achieved 83.72\%, while randomly reinitialized subnetworks achieved only 80.92\%, a 2.80 percentage point gap. This confirms that the original initialization contains information critical to the subnetwork's success, validating the core claim of the Lottery Ticket Hypothesis.

The slight improvement in accuracy observed at 70\% remaining parameters ($85.27\%$ vs.\ $84.87\%$ for the full model) suggests that moderate pruning may act as a form of implicit regularization, removing redundant or noisy parameters while preserving the essential learned representation.

\subsection{Relationships Among Mechanisms}

Taken together, the findings indicate that generalization in overparameterized neural networks cannot be attributed to a single mechanism. Instead, it appears to emerge from the interaction of several complementary factors:
\begin{enumerate}
    \item Optimization bias introduced by stochastic gradient descent biases training toward solutions with lower effective complexity.
    \item Loss landscape geometry favoring flat minima provides robustness to parameter perturbations and correlates with improved generalization.
    \item Model scaling effects associated with double descent demonstrate that increased capacity beyond the interpolation threshold can improve generalization rather than degrade it.
    \item Kernel-like behavior in wide networks provides stability during training, with wider networks requiring smaller parameter changes.
    \item Structural redundancy enables the discovery of sparse, well-performing subnetworks during training, consistent with the Lottery Ticket Hypothesis.
\end{enumerate}

These mechanisms are not independent. The implicit regularization of SGD appears to be the mechanism that guides optimization toward flat minima. The double descent phenomenon may be partially explained by the increasing availability of flat minima in larger models. And the Lottery Ticket Hypothesis suggests that overparameterization facilitates the discovery of favorable subnetworks precisely because larger search spaces contain more ``winning tickets.''

\subsection{Causal Hypotheses and Alternative Explanations}

While the preceding analysis identifies strong correlations between optimization dynamics, loss landscape geometry, and generalization performance, establishing definitive causal relationships requires careful consideration of alternative explanations and the limitations of the experimental evidence.

The experiments provide evidence consistent with a causal chain in which SGD's mini-batch noise causes convergence to flatter minima, which in turn causes improved generalization. However, this interpretation cannot be fully confirmed from observational experiments alone. An alternative hypothesis is that flat minima and good generalization are both consequences of a third factor, such as the alignment between the model's inductive bias and the data distribution's structure. The full-batch GD control partially addresses this concern---its simultaneous failure to reach flat minima and to generalize suggests that stochastic noise is at minimum a necessary condition for effective optimization in these architectures---but does not conclusively establish the causal direction between flatness and generalization.

Several specific alternative explanations can be evaluated against the experimental evidence. First, the hypothesis that generalization depends primarily on model capacity (as measured by parameter count) is inconsistent with the observation that identically-sized models trained with different batch sizes exhibit different generalization performance. Second, the hypothesis that generalization is driven entirely by explicit architectural choices (such as convolutional structure) is weakened by the observation that similar patterns hold for both MLPs and CNNs across different datasets. Third, the random reinitialization control in the lottery ticket experiments rules out the hypothesis that the pruned architecture alone---independent of initialization---accounts for subnetwork performance.

Several important questions remain unresolved by the present experiments. The precise mechanism by which mini-batch noise selects for flat minima over sharp ones is not fully characterized---whether this occurs through noise-driven escape from sharp basins, preferential attraction to wide basins, or a combination of both processes cannot be distinguished from the current data. The relationship between double descent and flat minima availability, while suggestive, lacks direct measurement of basin volume as a function of model scale, which would require loss landscape visualization techniques beyond the scope of this study. Finally, the extent to which these findings transfer to architectures with qualitatively different optimization dynamics, such as transformers with attention mechanisms and layer normalization, remains an open empirical question.

\section{Limitations}

Although the present study provides empirical and theoretical insight into the mechanisms underlying generalization in overparameterized neural networks, several limitations must be acknowledged.

\subsection{Dataset and Domain Limitations}

The experiments rely on MNIST and CIFAR-10, which are relatively constrained benchmarks. MNIST contains simple grayscale images, and CIFAR-10 contains small $32 \times 32$ natural images. The observed behaviors may not fully capture dynamics present in large-scale training scenarios involving transformers, large language models, or industrial-scale vision systems trained on datasets containing millions of samples.

\subsection{Architectural Scope}

The study focuses on multilayer perceptrons and convolutional neural networks with batch normalization. Modern systems increasingly rely on transformers, attention mechanisms, and large-scale foundation models with substantially different structural properties. The inclusion of batch normalization, while necessary for training stability, introduces a form of implicit regularization that may confound the isolation of optimization-induced regularization effects.

\subsection{Finite Experimental Scale}

Parameter counts in this study range from thousands to approximately 37 million. Contemporary production-scale models contain billions or trillions of parameters, and the observed phenomena may behave differently at those scales. The CIFAR-10 test accuracies (72--86\%) are below state-of-the-art, reflecting the deliberate exclusion of data augmentation and explicit regularization rather than limitations of the experimental framework.

\subsection{Loss Landscape Measurement Limitations}

The Hessian eigenvalue estimates are based on power iteration applied to a subset of training data and represent local approximations of curvature. Neural network loss landscapes exist in extremely high-dimensional parameter spaces, and the top eigenvalue captures only the direction of greatest curvature.

\subsection{NTK Regime Approximation}

Even the widest networks tested (width 4096) showed non-negligible parameter movement, indicating they had not fully converged to the NTK regime. True NTK behavior applies strictly in the infinite-width limit.

\section{Future Research Considerations}

Several promising directions emerge from this work:

Scaling experiments to large-scale architectures, particularly transformer-based language models and vision transformers, would test whether the observed mechanisms remain consistent at contemporary scales.

Developing unified theoretical frameworks that connect implicit regularization with loss landscape geometry and double descent behavior remains a major open challenge. The present study identifies empirical correlations between these phenomena but does not provide a formal mathematical framework unifying them.

Investigating the relationship between flat minima and adversarial robustness could clarify whether the same mechanisms that improve standard generalization also confer robustness to distributional shifts.

Extending lottery ticket analysis to identify whether winning tickets share structural properties across different initializations and architectures could provide deeper insight into the role of sparsity in neural network training.

\section{Conclusion}

This study investigated a central theoretical question in modern machine learning: why heavily overparameterized neural networks are able to generalize effectively despite possessing parameter counts that greatly exceed the number of training samples. Through controlled computational experiments on CIFAR-10 and MNIST, five complementary explanatory mechanisms were evaluated.

The experiments demonstrated that stochastic gradient descent introduces implicit regularization that biases training toward solutions with improved generalization, as evidenced by a monotonic 2.25 percentage point test accuracy improvement from batch size 2048 to batch size 32, with all models achieving identical training accuracy. Analysis of the loss landscape confirmed that better-generalizing solutions reside in flatter regions, with an $11.8\times$ difference in Hessian curvature between small-batch and large-batch trained models. The double descent phenomenon was observed in both MLP and CNN experiments, with test error continuing to decrease well beyond the interpolation threshold. Wide-network experiments showed progressive convergence toward NTK-predicted behavior, with relative parameter movement decreasing $11.3\times$ as width increased from 32 to 4096. Finally, pruning experiments confirmed the Lottery Ticket Hypothesis: sparse subnetworks retaining only 10\% of parameters achieved within 1.15 percentage points of full model accuracy when retrained from their original initialization, while randomly reinitialized subnetworks performed 2.80 percentage points worse.

These findings collectively demonstrate that parameter count alone does not adequately characterize the effective capacity of neural networks. Generalization depends on the interaction of architectural structure, optimization dynamics, loss landscape geometry, and the statistical properties of training data. While significant progress has been made in understanding individual components of the generalization puzzle, a fully unified theory of overparameterized learning remains an open challenge. Continued research integrating optimization theory, statistical learning principles, and high-dimensional geometry will be essential for explaining generalization in overparameterized models.

\bibliographystyle{IEEEtran}
\bibliography{references}

\appendix

\section{Experimental Hyperparameters}

The following table summarizes the primary hyperparameters used across experiments.

\begin{table}[htbp]
\centering
\caption{Default experimental hyperparameters.}
\label{tab:hyperparams}
\begin{tabular}{@{}ll@{}}
\toprule
Parameter & Value \\
\midrule
Optimizer                & SGD with momentum (default) \\
Momentum                 & 0.9 \\
Base Learning Rate       & 0.01 (at batch size 128) \\
LR Scaling               & Linear with batch size \\
Learning Rate Schedule   & Cosine annealing \\
Batch Size               & Varied (32--2048); 128 default \\
Weight Initialization    & PyTorch default (Kaiming/Xavier) \\
Activation Function      & ReLU \\
Training Epochs          & 60--80 \\
Weight Decay             & 0 (none, to isolate implicit reg) \\
Framework                & PyTorch 2.6.0 + CUDA 12.4 \\
GPU                      & NVIDIA GeForce RTX 3070 Ti (8 GB) \\
Random Seeds             & 5 per experiment \\
\bottomrule
\end{tabular}
\end{table}

For Adam optimizer comparisons:

\begin{table}[htbp]
\centering
\caption{Adam optimizer hyperparameters.}
\label{tab:adam_params}
\begin{tabular}{@{}ll@{}}
\toprule
Parameter & Value \\
\midrule
Optimizer & Adam \\
LR        & 0.001 \\
$\beta_1$ & 0.9 \\
$\beta_2$ & 0.999 \\
\bottomrule
\end{tabular}
\end{table}

\section{Network Architectures}

MLP architecture (variable width $w$, depth 4):
\begin{align*}
    &\text{Input} \to \text{Linear}(d_{\text{in}}, w) \to \text{ReLU} \\
    &\to \text{Linear}(w, w) \to \text{ReLU} \\
    &\to \text{Linear}(w, w) \to \text{ReLU} \\
    &\to \text{Linear}(w, 10)
\end{align*}
Width values: 8, 16, 32, 64, 128, 256, 512, 1024, 2048, 4096.

CNN architecture (variable base channels $c$). Each block consists of two Conv-BN-ReLU sequences followed by $2\times 2$ max pooling. All convolutions use $3\times 3$ kernels:
{\footnotesize
\begin{align*}
    &\text{Conv}(3,c) \!\to\! \text{BN} \!\to\! \text{ReLU} \!\to\! \text{Conv}(c,c) \!\to\! \text{BN} \!\to\! \text{ReLU} \!\to\! \text{Pool} \\
    &\text{Conv}(c,2c) \!\to\! \text{BN} \!\to\! \text{ReLU} \!\to\! \text{Conv}(2c,2c) \!\to\! \text{BN} \!\to\! \text{ReLU} \!\to\! \text{Pool} \\
    &\text{Conv}(2c,4c) \!\to\! \text{BN} \!\to\! \text{ReLU} \!\to\! \text{Conv}(4c,4c) \!\to\! \text{BN} \!\to\! \text{ReLU} \!\to\! \text{Pool} \\
    &\text{Flatten} \!\to\! \text{Linear}(64c, 256) \!\to\! \text{ReLU} \!\to\! \text{Linear}(256, 10)
\end{align*}}
Base channel values: 4, 8, 16, 32, 64, 96.

\section{Supplementary Mathematical Notes}

\subsection{Classical Generalization Bounds}

The expected generalization error can be bounded using the VC dimension $d_{\text{VC}}$:
\begin{equation}
    R(f) \leq R_{\text{emp}}(f) + O\!\left(\sqrt{\frac{d_{\text{VC}} \cdot \log(n / d_{\text{VC}})}{n}}\right)
\end{equation}
For networks with millions of parameters, $d_{\text{VC}}$ is extremely large, producing vacuous bounds.

\subsection{Interpolation Threshold}

The interpolation threshold occurs when model capacity $p$ satisfies $p \approx n$, where $n$ is the number of training samples. In the double descent framework, test error peaks near this threshold.

\subsection{Neural Tangent Kernel Formulation}

For a network $f(\mathbf{x}; \boldsymbol{\theta})$ with parameters $\boldsymbol{\theta}$, the NTK is defined as:
\begin{equation}
    K(\mathbf{x}, \mathbf{x}') = \left\langle \nabla_{\boldsymbol{\theta}} f(\mathbf{x}; \boldsymbol{\theta}),\; \nabla_{\boldsymbol{\theta}} f(\mathbf{x}'; \boldsymbol{\theta}) \right\rangle
\end{equation}
In the infinite-width limit, $K$ remains approximately constant during training, and the model prediction evolves as:
\begin{equation}
    \frac{d\mathbf{f}}{dt} = -K(\mathbf{X}, \mathbf{X})(\mathbf{f}(\mathbf{X}) - \mathbf{Y})
\end{equation}

\subsection{Flatness Metric}

The top Hessian eigenvalue $\lambda_{\max}$ provides a local measure of sharpness:
\begin{equation}
    \lambda_{\max} = \max_{\mathbf{v}} \frac{\mathbf{v}^\top \mathbf{H} \mathbf{v}}{\mathbf{v}^\top \mathbf{v}}
\end{equation}
where $\mathbf{H} = \nabla^2 \mathcal{L}(\boldsymbol{\theta})$ is the Hessian of the loss.

\subsection{Sparse Subnetwork Representation}

A pruned network is represented as $f(\mathbf{x}; \boldsymbol{\theta} \odot \mathbf{m})$, where $\mathbf{m} \in \{0,1\}^p$ is a binary mask and $\odot$ denotes elementwise multiplication. A ``winning ticket'' is a mask $\mathbf{m}$ such that the subnetwork trained from its original initialization $\boldsymbol{\theta}_0$ achieves performance comparable to the full model.

\section{Experimental Reproducibility}

Each experiment was repeated with 5 independent random seeds affecting weight initialization and mini-batch sampling order. Reported metrics represent the mean across runs, with standard deviations and 95\% bootstrap confidence intervals noted. Statistical comparisons use Welch's $t$-tests at the $\alpha = 0.05$ significance level.

\end{document}